%% file: paper_V1.tex
\documentclass[journal]{IEEEtran}
\usepackage{amsmath,graphicx}
\usepackage{cite}
\usepackage{xurl}
\usepackage{graphicx}
\usepackage{subcaption} 
\usepackage{caption}
\usepackage{amsfonts,amssymb,amsmath}
\usepackage{color}
 \usepackage{algorithm}
\usepackage{algorithmic}
 \usepackage{array}
 \usepackage{stfloats}
 \usepackage{float}
\usepackage{url}
 \usepackage{verbatim}
\usepackage{cite}
\usepackage{lipsum}
\usepackage{color}
\usepackage{tabularray}
\usepackage{adjustbox}
\usepackage{mathtools}
\usepackage[normalem]{ulem} 
\graphicspath{ {./imgs/} }
\usepackage[printonlyused,nohyperlinks]{acronym}
\usepackage[bookmarks=true,hidelinks,pdfpagelabels=true]{hyperref}
\usepackage{makecell}
\usepackage{booktabs}
\usepackage{fix-cm}

\usepackage{xcolor}
\definecolor{wildstrawberry}{rgb}{1.0, 0.26, 0.64}
\definecolor{update_green}{rgb}{0.4,0.8,0.0}
\definecolor{violet}{rgb}{0.55, 0, 0.55}
\definecolor{orange}{rgb}{1.0, 0.55, 0}

\title{
Bayesian Optimization applied for accelerated Virtual Validation of the Autonomous Driving Function }
\author{Satyesh Shanker Awasthi, Mohammed Irshadh Ismaaeel Sathyamangalam Imran, Stefano Arrigoni, and Francesco Braghin\thanks{All authors are with the Dipartimento di Meccanica, Politecnico di Milano, 20156 Milan, Italy.}}

\begin{document}

\maketitle
\input{Acronyms}
\begin{abstract}
Rigorous \acf{VV} of \acp{ADF} is paramount for ensuring the safety and public acceptance of \acp{AV}. Current validation relies heavily on simulation to achieve sufficient test coverage within the \ac{ODD} of a vehicle, but exhaustively exploring the vast parameter space of possible scenarios is computationally expensive and time-consuming. This work introduces a framework based on \acf{BO} to accelerate the discovery of critical scenarios. We demonstrate the effectiveness of the framework on an \ac{MPC}-based motion planner, showing that it identifies hazardous situations, such as off-road events, using orders of magnitude fewer simulations than brute-force \ac{DoE} methods. Furthermore, this study investigates the scalability of the framework in higher-dimensional parameter spaces and its ability to identify multiple, distinct critical regions within the \ac{ODD} of the motion planner used as the case study.

\end{abstract}


\input{Sections_V1/S1_Introduction}
\input{Sections_V1/S2_Related_Work}

\input{Sections_V1/S3_Methodology}

\input{Sections_V1/S4_UseCase_Validation_Simulations}

\input{Sections_V1/S5_Results}

\input{Sections_V1/S6_Conclusions_n_FutureWork}

\bibliographystyle{IEEEtran}
\bibliography{references_corrected}

\end{document}

%% file: Acronyms.tex
\acrodef{RSI}[RSI]{Roadside Infrastructure}
\acrodef{RSU}[RSU]{Roadside Unit}
\acrodef{SDK}[SDK]{Software Development Kit}
\acrodef{YOLOv4}[YOLOv4]{You Only Look Once v4}
\acrodef{DeepSORT}[DeepSORT]{Deep Simple Online and Realtime Tracking}
\acrodef{DL}[DL]{Deep Learning}
\acrodef{CNN}[CNN]{Convolutional Neural Network}
\acrodef{WRC}[WRC]{Weighted Residual Connections}
\acrodef{CSP}[CSP]{Cross Stage Partial}
\acrodef{CmBN}[CmBN]{Cross mini-Batch Normalization}
\acrodef{SAT}[SAT]{Self Adversarial Training}
\acrodef{SPP}[SPP]{Spatial Pyramid Pooling}
\acrodef{PAN}[PAN]{Path Aggregation Network}
\acrodef{SORT}[SORT]{Simple Online and Realtime Tracking}
\acrodef{MOT}[MOT]{Multi-Object Tracking}
\acrodef{VRU}[VRU]{Vulnerable Road User}
\acrodef{DNN}[DNN]{Deep Neural Network}
\acrodef{IoV}[IoV]{Internet of Vehicles}
\acrodef{MQTT}[MQTT]{Message Queuing Telemetry Transport}
\acrodef{MS COCO}[MS COCO]{Microsoft Common Objects in Context}
\acrodef{CIoU}[CIoU]{Complete Intersection over Union}
\acrodef{QoS}[QoS]{Quality of Service}
\acrodef{TLS}[TLS]{Transport Layer Security}
\acrodef{OAuth}[OAuth]{Open Authorization}
\acrodef{V2X}[V2X]{Vehicle-to-Everything}
\acrodef{AI}[AI]{Artificial Intelligence}
\acrodef{ETSI}[ETSI]{European Telecommunications Standard Institute}
\acrodef{OBU}[OBU]{On-Board Unit}
\acrodef{ROS}[ROS]{Robot Operating System}
\acrodef{CAN}[CAN]{Controller Area Network}
\acrodef{PoC}[PoC]{Proof of Concept}
\acrodef{L4T}[L4T]{Linux for Tegra}
\acrodef{TOPS}[TOPS]{Tera Operations Per Second}
\acrodef{BB}[BB]{Bounding Box}
\acrodefplural{BB}[BBs]{Bounding Boxes}
\acrodef{LLA}[LLA]{Latitude Longitude Altitude}
\acrodef{ECEF}[ECEF]{Earth-Centered Earth-Fixed}
\acrodef{ROI}[ROI]{Region Of Interest}
\acrodef{RTT}[RTT]{Round Trip Time}
\acrodef{ML}[ML]{Machine Learning}
\acrodef{CV}[CV]{Computer Vision}
\acrodef{4G}[4G]{4th Generation}
\acrodef{5G}[5G]{5th Generation}
\acrodef{6G}[6G]{6th Generation}
\acrodef{C-V2X}[C-V2X]{Cellular Vehicle-to-Everything}
\acrodef{VPN}[VPN]{Virtual Private Network}
\acrodef{FPS}[FPS]{Frames Per Second}
\acrodef{3GPP}[3GPP]{3rd Generation
Partnership Project}
\acrodef{LTE}[LTE]{Long Term Evolution}
\acrodef{BSM}[BSM]{Basic Safety Message}
\acrodef{CAM}[CAM]{Cooperative Awareness Message}
\acrodef{DENM}[DENM]{Decentralized Environmental Notification Message}
\acrodef{SL}[SL]{sidelink}
\acrodef{5G NR}[5G NR]{5G New Radio}
\acrodef{URLLC}[URLLC]{Ultra-Reliable Low Latency Communications}
\acrodef{FR2}[FR2]{Frequency Range 2}
\acrodef{NR}[NR]{New Radio}
\acrodef{WHO}[WHO]{World Health Organization}
\acrodef{CAV}[CAV]{Connected and Autonomous Vehicle}
\acrodef{NHTSA}[NHTSA]{National Highway Traffic Safety Administration}
\acrodef{V2V}[V2V]{Vehicle-to-Vehicle}
\acrodef{V2I}[V2I]{Vehicle-to-Infrastructure}
\acrodef{V2P}[V2P]{Vehicle-to-Person}
\acrodef{AEB}[AEB]{Autonomous Emergency Braking}
\acrodef{ADAS}[ADAS]{Advanced Driver Assistance Systems}
\acrodef{CCA}[CCA]{Cooperative Collision Avoidance}
\acrodef{CPM}[CPM]{Collaborative Perception Message}
\acrodef{TTR}[TTR]{Time To Reach}
\acrodef{TTC}[TTC]{Time to Collision}
\acrodef{FCW}[FCW]{Forward Collision Warning}
\acrodef{CWAS}[CWAS]{Collision Warning and Avoidance System}
\acrodef{P-ICWAS}[P-ICWAS]{Pedestrian-Intersection Collision Warning and Avoidance System}
\acrodef{HMI}[HMI]{Human Machine Interface}
\acrodef{UTM}[UTM]{Universal Transverse Mercator}
\acrodef{UKF}[UKF]{Unscented Kalman Filter}
\acrodef{GPS}[GPS]{Global Positioning System}
\acrodef{RTK}[RTK]{Real-Time Kinematic}
\acrodef{ECU}[ECU]{Electronic Control Unit}
\acrodef{WLAN}[WLAN]{Wireless Local Area Network}
\acrodef{DSRC}[DSRC]{Dedicated Short-Range Communication}
\acrodef{FCC}[FCC]{Federal Communications Commission}
\acrodef{MAC}[MAC]{Medium Access Control}
\acrodef{PHY}[PHY]{Physical Layer}
\acrodef{WAVE}[WAVE]{Wireless Access in Vehicular Environment}
\acrodef{C-ITS}[C-ITS]{Cooperative Intelligent Transport System}
\acrodef{ITS}[ITS]{Intelligent Transport System}
\acrodef{FR}[FR]{Frequency Range}
\acrodef{RAT}[RAT]{Radio Access Technology}
\acrodef{SPS}[SPS]{Semi-Persisten Scheduling}
\acrodef{CAM}[CAM]{Cooperative Awareness Messages}
\acrodef{BSM}[BSM]{Basic Safety Messages}
\acrodef{eNB}[eNB]{evolved Node-B}
\acrodef{eV2X}[eV2X]{enhanced V2X}
\acrodef{5GAA}[5GAA]{5G Automotive Association}
\acrodef{GLOSA}[GLOSA]{Green Light Optimal Speed Advisory}
\acrodef{mmWave}[mmWave]{Millimeter wave}
\acrodef{HT-URLLC}[HT-URLLC]{High Throughput
Ultra-Reliable Low Latency Communication}
\acrodef{KPI}[KPI]{Key Performance Indicator}
\acrodef{LiDAR}[LiDAR]{Light Detection And Ranging}
\acrodef{RGB}[RGB]{Red, Green, and Blue}
\acrodef{MEC}[MEC]{Multi-access Edge Computing}
\acrodef{JSON}[JSON]{JavaScript Object Notation}
\acrodef{BEV}[BEV]{Bird's Eye View}
\acrodef{GNSS}[GNSS]{Global Navigation Satellite System}
\acrodef{IMU}[IMU]{Inertial Measurement Unit}
\acrodef{SLAM}[SLAM]{Simultaneous Localization And Mapping}
\acrodef{EKF}[EKF]{Extended Kalman Filter}
\acrodef{RMSE}[RMSE]{Root Mean Square Error}
\acrodef{FOV}[FOV]{Field Of View}
\acrodef{IoT}[IoT]{Internet of Things}
\acrodef{LOS}[LOS]{Line-of-Sight}
\acrodef{CDF}[CDF]{Cumulative Distribution Function}
\acrodef{ADF}[ADF]{Autonomous Driving Function}
\acrodef{HAV}[HAV]{Highly Automated Vehicle}
\acrodef{ODD}[ODD]{Operational Design Domain}
\acrodef{AV}[AV]{Autonomous Vehicle}
\acrodef{MOT}[MOT]{Multi-Object Tracking}
\acrodef{VRU}[VRU]{Vulnerable Road User}
\acrodef{V2X}[V2X]{Vehicle-to-Everything}
\acrodef{OBU}[OBU]{On-Board Unit}
\acrodef{TTR}[TTR]{Time To Reach}
\acrodef{TTC}[TTC]{Time to Collision}
\acrodef{UKF}[UKF]{Unscented Kalman filter}
\acrodef{GPS}[GPS]{Global Positioning System}
\acrodef{RTK}[RTK]{Real-Time Kinematic}
\acrodef{KPI}[KPI]{Key Performance Indicator}
\acrodef{LiDAR}[LiDAR]{Light Detection And Ranging}
\acrodef{BEV}[BEV]{Bird's Eye View}
\acrodef{GNSS}[GNSS]{Global Navigation Satellite System}
\acrodef{IMU}[IMU]{Inertial Measurement Unit}
\acrodef{SLAM}[SLAM]{Simultaneous Localization And Mapping}
\acrodef{EKF}[EKF]{Extended Kalman Filter}
\acrodef{RMSE}[RMSE]{Root Mean Square Error}
\acrodef{FOV}[FOV]{Field Of View}
\acrodef{IoT}[IoT]{Internet of Things}
\acrodef{OCP}[OCP]{Optimal Control Problem}
\acrodef{CBF}[CBF]{Control Barrier Function}
\acrodef{OV}[OV]{Obstacle Vehicle}
\acrodef{CLF}[CLF]{Control Lyapunov Function}
\acrodef{QP}[QP]{Quadratic Programming}
\acrodef{APF}[APF]{Artificial Potential Field}
\acrodef{MPC}[MPC]{Model Predictive Controller}
\acrodef{STM}[STM]{Single Track Model}
\acrodef{DoF}[DoF]{Degree of Freedom}
\acrodef{DoE}[DoE]{Design of Experiments}
\acrodef{CoG}[CoG]{Center of Gravity}
\acrodef{VVUQ}[VVUQ]{ Verification, Validation and Uncertainty Quantification}
\acrodef{VV}[V\&V]{ Verification and Validation}
\acrodef{UQ}[UQ]{Uncertainty Quantification}
\acrodef{ODD}[ODD]{Operational Design Domain}
\acrodef{PET}[PET]{Post Encroachment Time}
\acrodef{AD}[AD]{Autonomous Driving}
\acrodef{HAD}[HAD]{Highly Automated Driving}
\acrodef{ALKS}[ALKS]{Automated Lane Keeping Systems}
\acrodef{ADS}[ADS]{Autonomous Driving Systems}
\acrodef{ACC}[ACC]{Adaptive Cruise Control}
\acrodef{SAE}[SAE]{SAE}
\acrodef{ASAM}[ASAM]{Association for Standardization of Automation and Measuring Systems}
\acrodef{SiL}[SiL]{Software-in-the-Loop}
\acrodef{MiL}[MiL]{Model-in-the-Loop}
\acrodef{HiL}[HiL]{Hardware-in-the-Loop}
\acrodef{XiL}[XiL]{X-in-the-Loop}
\acrodef{SOTIF}[SOTIF]{Safety of the Intended Functionality}
\acrodef{ADTF}[ADTF]{Automotive Data and Time-Triggered Framework }
\acrodef{HPC}[HPC]{High-Performance Computing}
\acrodef{RL}[RL]{Reinforcement Learning}
\acrodef{BO}[BO]{Bayesian Optimization}
\acrodef{DBSCAN}[DBSCAN]{Density-Based Spatial Clustering of Applications with Noise}

%% file: Sections_V1/S1_Introduction.tex
\section{Introduction}
\label{sec:1}
\subsection{Why virtual testing and validation is necessary?}
\acfp{AV} promise to revolutionize urban and highway mobility throughout the world. However, the reliability of autonomous driving needs to be proven sufficiently and in little time as possible. On-road testing of \acfp{HAV} alone is not sufficient to prove the reliability of \ac{ADS}, as high-risk driving situations are unlikely to be encountered randomly in real-world miles obtained from naturalistic driving~\cite{KALRA2016182,edgeCorner}. The space of possible scenarios is effectively infinite, making exhaustive testing infeasible~\cite{Rev_Crit}. Moreover, experimental testing might not even be possible in the early \ac{VV} phase of \acp{ADS} or might be time-consuming and costly~\cite{challenges_vv}. Thus, simulation-based virtual testing is essential for AV safety validation~\cite{survey_virtualTesting,statistics}. \\

\subsection{Regulations regarding virtual validation}
In recent years, international and regional regulations have recognized the reasons stated above and have initiated accommodating and standardizing, testing and certification of such vehicles. The World Forum for Harmonization of Vehicle Regulations (WP.29) drafts define “virtual testing” and allow limited use of simulation in type approval~\cite{unece}. The UNECE-R157 regulation for \ac{ALKS} already permit certain \ac{SiL}/\ac{HiL} tests for variant approval~\cite{unece}. \ac{ASAM} developed a holistic best practice blueprint to exemplify the diverse testing strategies~\cite{ASAM_testSpec}. In practice, companies like TÜV SÜD advocate a modular validation pipeline by mixing \ac{MiL}, \ac{SiL}, and \ac{HiL} tests and systematic scenario catalogs for type approval~\cite{xil,xil1,xil2}. 
\subsection{Fallbacks of virtual testing}
However, even simulation-based testing can be computationally intensive. Brute-force methods like combinatorial testing and Monte Carlo simulation typically require a prohibitively large number of runs for confidence~\cite{surveySBT,GAS}. Advanced uncertainty quantification techniques account for variations in environmental conditions, sensor noise, and behavioral uncertainties, providing robust statistical foundations for safety claims~\cite{EnvSensNoise,statistics}. Sufficient scenario coverage can lead to a significant effort in simulation time and cost and is directly proportional to the number of parameters chosen.
To overcome these limitations, generally, two solutions have been explored. The first strategy involves increasing the computational power by implementing massive parallelization of the simulations~\cite{waymax,parallel}. The second and more efficient option reduces the number of simulation by using scenario-based frameworks for expediting the scenario search for higher risk scenarios~\cite{surveySBT,Rev_Crit}. These tests result in the critical scenarios which could then be retested on the proving grounds for higher fidelity~\cite{xil2}.  While testing only the most critical situations is sufficient for 'type approving' established systems, a different approach is needed for self-driving systems that are still in development. For these developing systems, safety and performance verification must involve continuous and rapid simulations of both everyday driving situations and high-risk edge cases. These solutions are discussed more in detail in Section~\ref{sec:2}.
\subsection{Contributions}
 This paper presents a virtual testing and homologation framework in which the critical scenario discovery is formulated as a \acf{BO} problem. This general validation workflow can be applied to any logical scenario with a given number of input dimensions within a defined Operational Design Domain (ODD). The performance of the framework was further demonstrated in a case study on an autonomous driving motion planner previously developed by the authors~\cite{Rule_def_MPC}. 
The main contributions of this paper are as follows:
\begin{enumerate}
 \item \acf{BO} framework: We formulate critical scenario discovery as a \acf{BO} problem to find scenario parameters that maximize a safety-violation metric. 
The framework is agnostic to the scenario specification, allowing any logical scenario with continuous parameters within an \ac{ODD} to be directly integrated into the Bayesian Optimization loop. 
 \item Application on a developed algorithm: The works present in the literature either focus on early detection of criticality from standard traffic simulations or they are based on naturalistic datasets. To the best knowledge of the authors, this is the first work demonstrating the use of such a framework on a developed \ac{ADF}.
 \item Scalability and multiple critical cluster identification: As the number of input dimensions increases, the method consistently identifies failure clusters using one to two orders of magnitude fewer simulations than the equivalent brute-force, full-factorial baseline. Furthermore, the framework successfully locates failures across different types of critical zones that were initially identified through the comprehensive baseline analysis in the applied case study.
\end{enumerate}
 \subsection{Arrangement of the paper}
The rest of this paper is organized as follows. Section~\ref{sec:2} reviews work related to acceleration of virtual testing techniques. Section~\ref{sec:3} discusses the developed framework in detail, including the optimization problem formulation. Section~\ref{sec:4} describes the co-simulation methodology as well as the autonomous driving function used to validate the framework. Section~\ref{sec:5} outlines the outcomes and results of the simulations. Finally, Section~\ref{sec:6} concludes and suggests future work.

%% file: Sections_V1/S2_Related_Work.tex
\section{Related Work}
\label{sec:2}
 A \acf{VV} approach using scenario-based testing for \acp{HAV} within a highway \ac{ODD} was defined in the PEGASUS project~\cite{pegasus}. Scenarios within this \ac{ODD} were classified into functional, logical, and concrete levels to build a comprehensive simulation catalog~\cite{pegasus2}. This methodology uses extensive simulation to identify critical parameter regions, which are subsequently validated on test tracks~\cite{pegasus3}. Follow-on projects, such as V\&V-Method and SETLevel, have since expanded the \ac{ODD} to include urban intersection scenarios~\cite{VVM,SETLevel}.

Standard random or combinatorial testing is often inefficient at identifying rare failure events. Given that critical scenarios are rare, prioritizing their discovery is essential for reducing testing redundancy. Consequently, recent research has increasingly focused on learning-based and optimization-driven methods to rapidly converge on these safety-critical scenarios. Several prominent approaches from recent literature are discussed below. 
One prominent and obvious strategy involves parallel and distributed simulation. In the field of mobility testing, solutions to simulate thousands of scenarios simultaneously, achieving significant improvements in computational speed have been explored. For instance, \cite{HPC}~presents an \ac{HPC} pipeline that executes Webots and SUMO simulations in parallel by distributing batch jobs across multiple nodes and instances. In addition, new simulators accelerated by TPUs and GPUs, such as JAX-based engines, are emerging. The Waymax simulator is a differentiable GPU-based traffic simulator capable of simulating up to 128~agents in parallel~\cite{waymax}. These strategies have not been explored in the present work but might be combined with these methods to exploit such simulation toolchains. However, specialized hardware is expensive to purchase and maintain, and keeping up with the latest versions can add significant costs. Furthermore, not all simulation software can utilize hardware acceleration, limiting its benefits and requiring specialized expertise to optimize code for heterogeneous hardware environments.\\
The emergence of cloud-based simulation platforms represents a shift towards distributed and scalable autonomous vehicle testing. NVIDIA DRIVE Constellation exemplifies the approach through its dual-server architecture, where one server runs DRIVE Sim software to simulate sensor data while a second server contains the DRIVE Pegasus AI car computer running the complete autonomous vehicle software stack~\cite{nvidia_pegasus}. The AVxcelerate Autonomy platform demonstrates the potential of robust, adaptive exploration algorithms optimized for cloud environments, claiming to accelerate sensitivity analyses by a magnitude of $10^3$ compared to traditional algorithms~\cite{ansys_avxcelerate}. These solutions too could eventually be used in conjunction with the work presented in the paper and have not been explored further. Nevertheless, large-scale cloud usage can become costly over time and moving large simulation datasets to and from the cloud can be slow and expensive.\\
A more targeted approach uses learning-based techniques to actively generate critical scenarios. \Ac{RL} methods, for example, train agents to discover safety violations by modifying scenario parameters like agent trajectories~\cite{MOEQT,d2rl,RL_sbt}. DynNPC trains non-player characters to provoke ego-vehicle failures, identifying them 92\% faster than static tests~\cite{DynNPC}. The main objective of these methods is not scenario-coverage but simulating maximum violations achieved during a each test run as quickly as possible after getting trained on a significant amount of episodes. This adversarial approach needs to be re-learned each time the ego vehicle or the underlying \acf{ADF} changes, rendering this method inefficient for development or homologation processes. \ac{RL} requires extensive simulation episodes to converge, and shaping reward functions to produce realistic yet critical scenarios is non-trivial. If improperly constrained, these methods can generate physically implausible or irrelevant failure cases.

To mitigate the high computational cost of simulation,~\cite{GAS} employs surrogate modeling on the whole vehicle system. Since high-fidelity \ac{AV} simulations are the primary bottleneck, surrogate models built from cheaper, approximate models can accelerate evaluation. However, surrogates are only as good as their training data. Their fidelity can degrade if the operational domain shifts, and they may struggle to capture the highly nonlinear dynamics of complex perception and control systems. 
Instead of using the surrogates directly for obtaining the results of the simulation, optimization-based testing uses iterative optimization on surrogate models to guide the search toward failure regions efficiently~\cite{surr1,surr2}. Heuristic-based optimization methods like genetic algorithms for test parameter optimization for \ac{ADAS} testing have also been explored~\cite{Rev_Crit,gen_opt,dec_tree}. Bayesian Optimization has also been explored in literature to identify parameterized test conditions that can lead to system failure and to generate new scenarios to be tested in urban scenarios in~\cite{BO1,vvm2}
A prior work has studied the use of \ac{BO} with different types of surrogate models to find the worst-case failures in a determined parameter space~\cite{BO2}. This paper leverages Bayesian Optimization to develop a framework for testing and \ac{VV} of autonomous vehicles. This framework is then applied to an example autonomous driving function to study the scalability and discovery of multiple critical zones in the ODD of the driving function.


%% file: Sections_V1/S3_Methodology.tex
\section{Problem Formulation : Framework}
\label{sec:3}
The primary goal of this paper is to introduce a framework for the rapid testing and validation of \acfp{ADF}, saving significant time and resources during both development and homologation. To achieve this, we have formulated a framework that utilizes Bayesian Optimization (BO) to maximize the probability of simulating critical scenarios. BO is a sample-efficient global optimization method particularly well-suited for problems involving expensive or black-box objective functions, such as high-fidelity vehicle simulations. In the context of scenario-based validation, BO efficiently explores the vast space of simulation parameters to identify critical scenarios—those leading to near-failure or safety-critical behavior—with a minimal number of simulation runs. The process works by modeling an underlying objective, such as a safety metric, with a probabilistic surrogate model like a Gaussian Process. An acquisition function then sequentially selects new test cases by intelligently balancing exploration of uncertain parameter regions and exploitation of areas already known to be critical. This methodology accelerates the discovery of edge cases within the operational design domain (ODD).
This approach involves parametric modification of key environmental and behavioral variables, including weather conditions, traffic density, road geometry, and agent behaviors~\cite{6layer}. The critical scenarios are discovered efficiently while maintaining the traceability and repeatability of test conditions.
\subsection{Problem Formulation}
The logical scenario ($S$) is defined as a set of continuous parameters $x \in \mathbb{R}^n$
that describe $n$ number of initial, environment, and dynamic parameters with the \ac{ODD} of the vehicle. Given $x$, the simulation produces a failure or criticality metric $f(x)$ that quantifies the criticality of the scenario. Thus, the problem of finding the critical scenario reduces to identifying $x$ that maximizes $f(x)$. Since each evaluation of $f(x)$ requires an expensive simulation run, this is cast as a black-box optimization problem.
\subsection{The Optimization framework}
\begin{figure*}[t] 
  \centering
  \includegraphics[width=\textwidth]{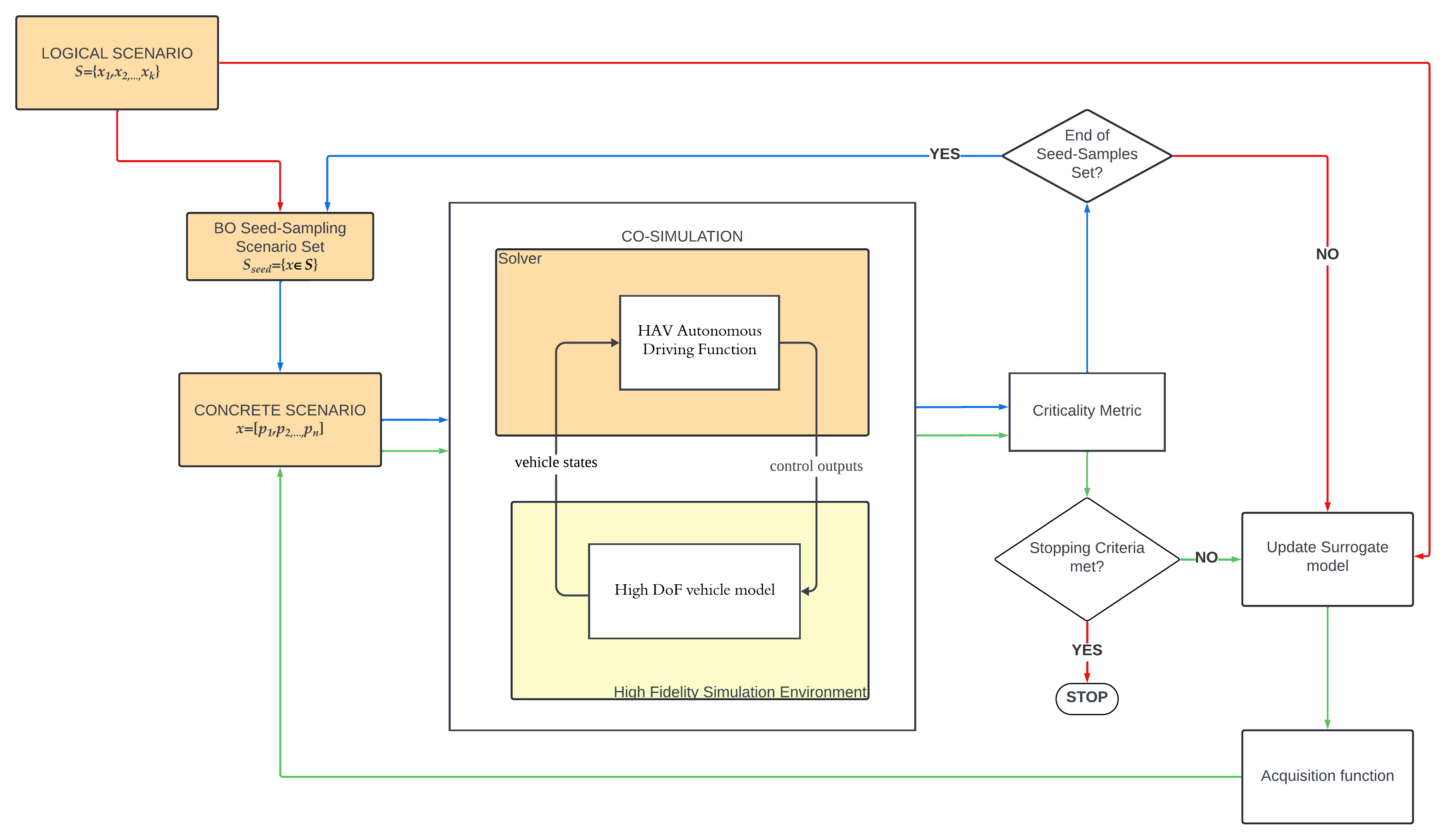}
  \caption{Layout of feedback loop established between the developed motion planner and the high-fidelity simulation software in the accelerated validation framework loop.}
  \label{fig:arch}
\end{figure*}

Figure~\ref{fig:arch} illustrates the overall architecture of the proposed simulation acceleration framework using Bayesian Optimization (BO) for critical scenario discovery in autonomous vehicle testing. The process begins with the definition of a logical scenario ($S$), where the operational design domain (ODD) is parameterized (e.g., ego speed, obstacle speed, initial ego position), $x=[p_1,p_2,...,p_n]$. A concrete scenario is instantiated by sampling a specific set of parameters within these ranges. Initial seed samples ($S_{seed}$) are selected to randomly initialize the BO process as represented by the blue arrow loop. Each sampled scenario is evaluated through a co-simulation environment, which couples a high-fidelity vehicle dynamics simulator and a partial or full \ac{ADF} implementation. This environment simulates the autonomous driving function in a realistic context, producing vehicle state trajectories and control signals. A criticality metric such as \ac{TTC} or \ac{PET} is computed for each simulation to assess the criticality of the simulation for the given parameter set~\cite{criticality_metrics}. Based on the outcomes, a surrogate model is updated and used by an acquisition function to propose the next promising scenario to evaluate as represented by the green arrow loop in Figure~\ref{fig:arch}. The loop continues until a stopping criterion is met, for example, a maximum number of simulations is reached. In this work, the loop terminates either when a decided number of simulations have been completed or until at least one instance in all the critical clusters, identified during baseline simulations, has been found. This architecture ensures that computational resources are focused on discovering high-risk scenarios efficiently.

\begin{algorithm}[!t]
\caption{Bayesian Optimization}
\label{alg:bayesopt}
\begin{algorithmic}[1]
\REQUIRE Objective function: Criticality metric $f(x)$, search domain: parameter ranges $\mathcal{X}$, initial dataset: Seed Samples $\mathcal{D}_0 = \{(x_i, f(x_i))\}_{i=1}^{n_0}$, acquisition function $\alpha(x;\mathcal{D})$, maximum iterations or stopping criteria $T$
\ENSURE Approximate maximizer $x^*$
\STATE Initialize surrogate model (e.g., Gaussian Process) with $\mathcal{D}_0$
\FOR{$t = n_0 + 1$ to $T$}
    \STATE Select next point $x_t = \arg\max_{x \in \mathcal{X}} \alpha(x;\mathcal{D}_{t-1})$
    \STATE Evaluate objective: $y_t = f(x_t)$
    \STATE Update dataset: $\mathcal{D}_t = \mathcal{D}_{t-1} \cup \{(x_t, y_t)\}$
    \STATE Update surrogate model with $\mathcal{D}_t$
\ENDFOR
\STATE \textbf{return} $x^* = \arg\max_{x_i \in \mathcal{D}_T} f(x_i)$
\end{algorithmic}
\end{algorithm}

The algorithmic workflow is presented in Algorithm~\ref{alg:bayesopt}. The description of individual components of the working loop is as following:
\begin{enumerate}
\item  \textit{Initialization}: Sample an initial set of scenario parameters $\{x_i\}$ and evaluate the simulator to obtain $\{f(x_i)\}$. The choice of sampling method and number of initial points affects convergence speed and overall performance. There are several methods to choose these initial set of simulations  such as: grid sampling, random sampling, sobol sequence and latin hypercube sampling. Any of these methods may be used in this framework but latin hypercube sampling has been used in the case study presented later because it promises better input space coverage~\cite{LHS}. The initial samples increase with the dimensionality of the logical scenario parameters. As a rule of thumb, the seed samples may range from two to five times the number of parameters chosen~\cite{frazier_tut}.
\item  \textit{Surrogate model}: Fit a Gaussian process (GP) regression model to the accumulated data $\{(x_i, f(x_i))\}$. The GP provides a posterior mean and variance over the function $f(x)$. Other probabilistic surrogate models include bayesian neural networks and tree-based random forest.
In this work, the underlying objective function $f(x)$, representing the criticality metric in a scenario defined by parameters $x \in \mathbb{R}^n$, is modeled as a Gaussian Process (GP):

\begin{equation}
f(x) \sim \mathcal{GP}\left(0, k(x, x')\right),
\end{equation}

where $k(x, x')$ is the covariance function or kernel encoding assumptions about the smoothness of the function. It defines the covariance between the values of the function at two different input points, $x$ and $x'$. The surrogate model is a zero-mean Gaussian Process with a squared exponential kernel given by:

\begin{equation}
k_{\text{SE}}(x, x') = \sigma_f^2 \exp\left(-\frac{1}{2}(x - x')^\top \Lambda^{-1} (x - x')\right),
\end{equation}

where $\Lambda = \text{diag}(\ell_1^2, \ell_2^2, \ldots, \ell_d^2)$ is a diagonal matrix of squared length-scales, and $\sigma_f^2$ is the signal variance.

Hyperparameters of $f(x)$ (length-scales and variance) are learned by maximizing the marginal likelihood. The hyperparameters $\theta = \{\ell_i, \sigma_f^2, \sigma_n^2\}$ are estimated by maximizing the log marginal likelihood:

\begin{equation}
\log p(\mathbf{y} \mid \mathbf{X}, \theta) = -\frac{1}{2} \mathbf{y}^\top K_\theta^{-1} \mathbf{y} - \frac{1}{2} \log |K_\theta| - \frac{n}{2} \log 2\pi,
\end{equation}

where $K_\theta = K + \sigma_n^2 I$ is the covariance matrix with added observation noise $\sigma_n^2$ and $\textbf{y}$ is the observed data at each $x$ in $\textbf{X}$.

The GP surrogate captures the relationship between scenario parameters and the criticality metric. We rely on the GP surrogate to estimate which untested scenarios may be critical. Given training data $\mathcal{D} = \{ \mathbf{X}, \mathbf{y} \}$, the predictive distribution at a new input $x_*$ is Gaussian:

\begin{align}
\mu(x_*) &= k_*^\top K^{-1} \mathbf{y}, \\
\sigma^2(x_*) &= k(x_*, x_*) - k_*^\top K^{-1} k_*,
\end{align}

where $k_* = k(\mathbf{X}, x_*)$ is the covariance vector between $x_*$ and the training points.
\item   \textit{Acquisition}: After the GP posterior has been evaluated, the next input point for the simulation is chosen using an acquisition function, which guides the search by balancing exploration and exploitation. In this work, we use Thompson Sampling, a Bayesian strategy where a random function, $\tilde{f}(x)$, is drawn from the GP posterior. The next point to simulate, $x_{\text{next}}$, is selected by finding the maximizer of this sampled function, as if it were the true objective:
\begin{equation} 
    x_{\text{next}} = \arg\max_{x \in \mathcal{X}} \tilde{f}(x)
    \label{eq:x_next}
\end{equation}
where $\mathcal{D} = \{(x_i, y_i)\}_{i=1}^{n}$ is the current dataset. In Section~\ref{sec:5}, this method is compared to the \textit{Probability of Improvement} acquisition function, which is a utility-based criterion defined as the likelihood that a new candidate point $x$ will improve over the current best observed value $y^{+}$ (assuming maximization):

\begin{equation}
    \text{PI}(x) = \mathbb{P}(f(x) \geq y^{+} + \xi) = \Phi\left( \frac{\mu(x) - y^{+} - \xi}{\sigma(x)} \right),
\end{equation}

where:
\begin{itemize}
    \item $\mu(x)$ and $\sigma(x)$ are the GP posterior mean and standard deviation,
    \item $y^{+} = \max_{i=1,\ldots,n} y_i$ is the best observed value,
    \item $\xi \geq 0$ is a margin parameter encouraging exploration,
    \item $\Phi(\cdot)$ is the CDF of the standard normal distribution.
\end{itemize}

The next point to be simulated, as in the case of Equation~\ref{eq:x_next}, when using probability of improvement acquisition function is selected by:
\begin{equation}
    x_{\text{next}} = \arg\max_{x \in \mathcal{X}} \text{PI}(x)
\end{equation}
Other acquisition functions like expected Improvement and Upper Confidence Bound have been tested too but not included in the results presented.
\item \textit{Evaluation}: Run the simulator at $x_{\text{next}}$ and compute  $f(x_{\text{next}})$.
\item  \textit{Update}: Augment the dataset with $(x_{\text{next}}, f(x_{\text{next}}))$
 and go to step 2.

\end{enumerate}
The process repeats until a budget of simulations is exhausted or a convergence criterion is met. BO builds a surrogate for the objective with a GP and uses an acquisition function to decide where to sample.


%% file: Sections_V1/S4_UseCase_Validation_Simulations.tex
\section{Framework Validation Setup: Methodology}
\label{sec:4}
\begin{figure*}[b] 
  \centering

  \begin{subfigure}[b]{0.5\textwidth}
    \centering
    \includegraphics[width=\textwidth]{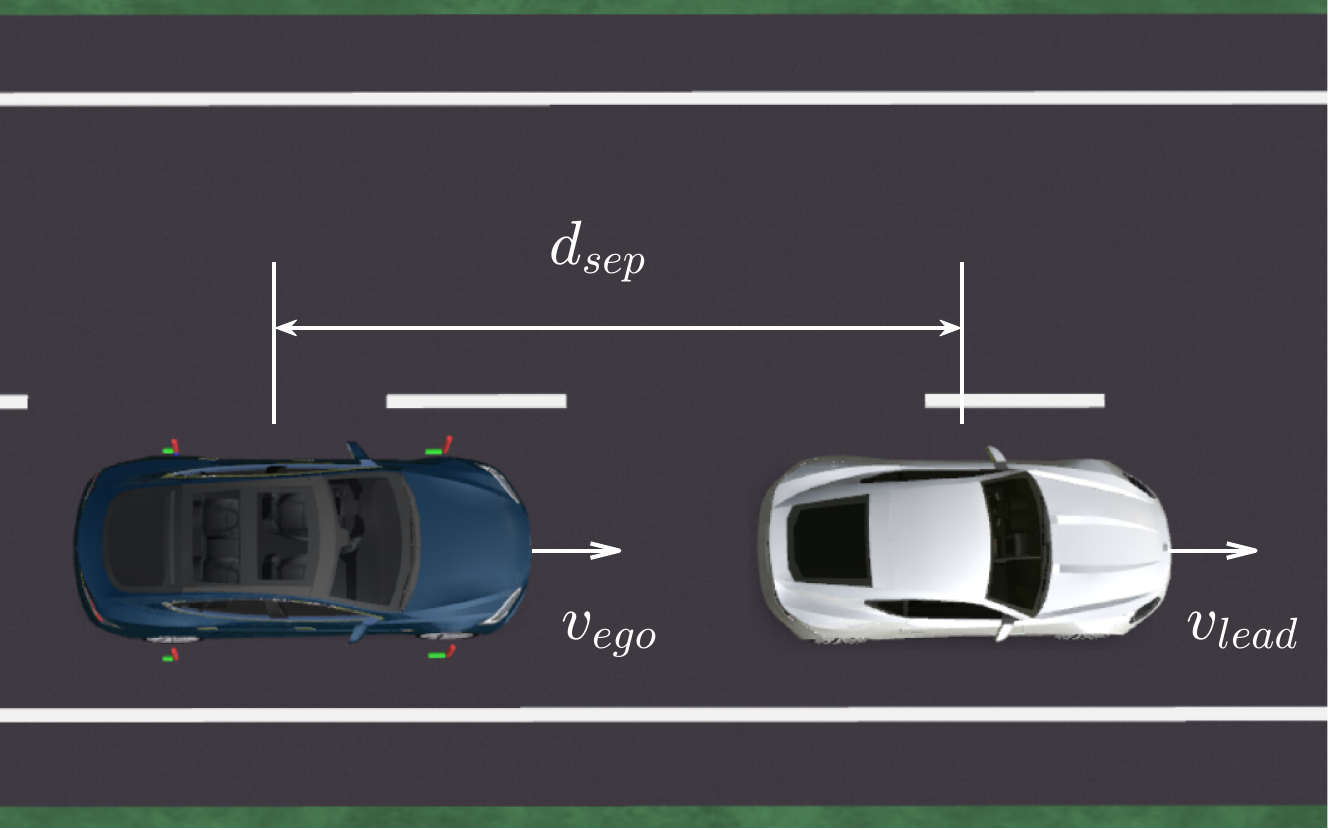}
    \caption{3D scenario}
    \label{fig:3d}
  \end{subfigure}
  \hfill
  \begin{subfigure}[b]{0.45\textwidth}
    \centering
    \includegraphics[width=\textwidth]{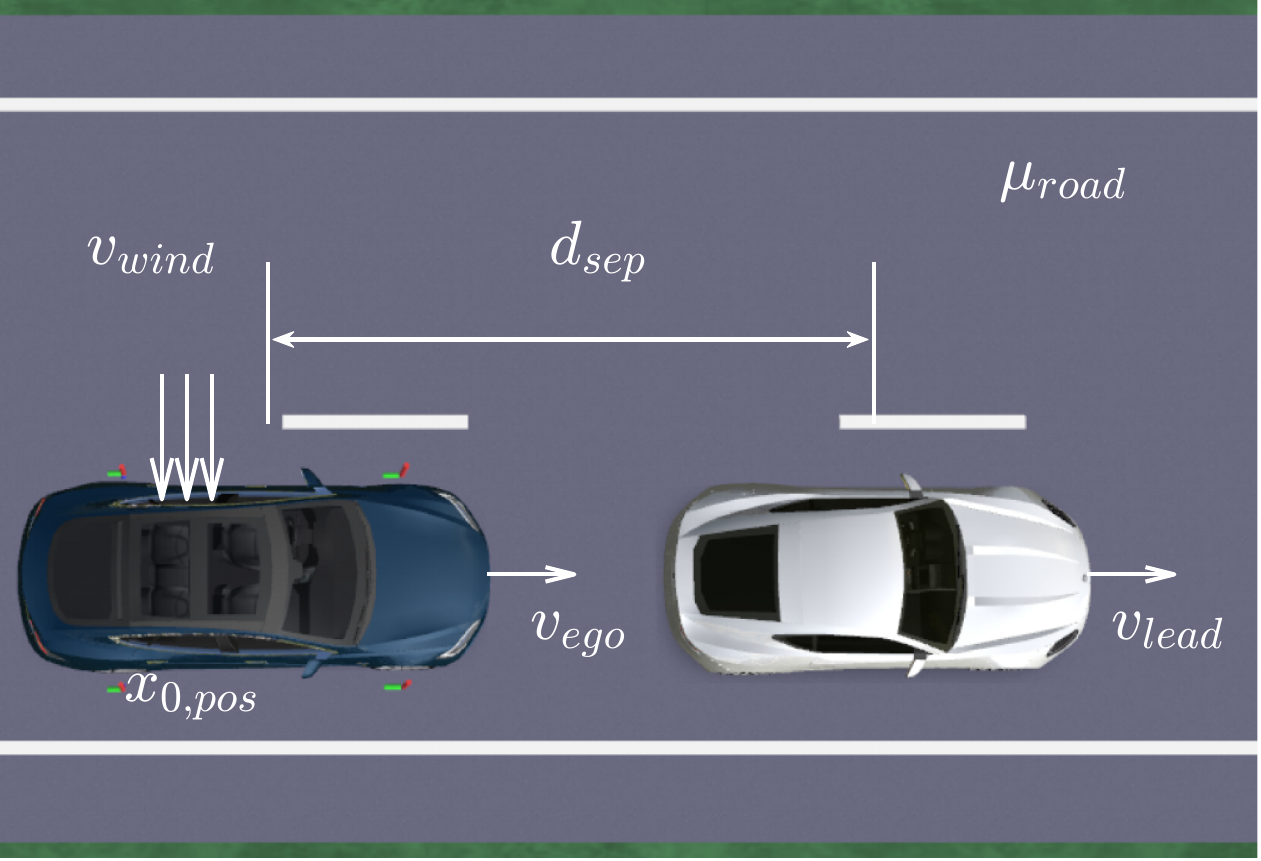}
    \caption{6D scenario}
    \label{fig:6d}
  \end{subfigure}
 
  \caption{A scene from the logical scenario being tested and three variable parameters of the logical scenario}
  \label{fig:scenarios}
\end{figure*}

In this section, the virtual validation framework is applied to a specific use case: a motion planner for a Highly Automated Vehicle (HAV) operating within an Operational Design Domain (ODD) defined as a straight, two-lane highway with preceding traffic. It should be noted that the \ac{ODD} in which this example \ac{ADF} works has been oversimplified for demonstration purposes. The motion planner follows a straight path and avoids the obstacles that fall into its path by overtaking or following. A detailed overview of the motion-planner formulation is presented in~\cite{Rule_def_MPC}. The ego vehicle starts in the right lane behind the obstacle vehicle. The aim of the study is to validate that the developed framework finds the critical scenarios or failures for the algorithm being tested faster than the baseline methods. The results obtained from the validation framework are compared with the validation results obtained from baseline combinatorial simulations. In addition to the general methodology discussed in Section~\ref{sec:3}, the choice of various components in Figure~\ref{fig:arch} for this particular application has also been discussed in detail. 
\\
To demonstrate the intuition behind the validation method and then the scalability of the parameter space, two sets of parameters have been chosen. Firstly, a set of 3 variable input parameters (3-\ac{DoF} simulation) is chosen: $x_{\text{3-DoF}}=[v_{\text{ego}}, v_{\text{lead}}, d_{\text{0,sep}}]$, which consists of ego-vehicle velocity, leading vehicle velocity, and initial distance between the two vehicles as shown in Figure~\ref{fig:3d}. A deeper investigation is then provided extending the original set with other 3 additional parameters (6-\ac{DoF} simulations) that are: lateral ego vehicle position along the lane, wind-gust velocity and road friction representing the environmental parameters i.e. $x_{\text{6-DoF}}=[v_{\text{ego}}, v_{\text{lead}}, d_{\text{0,sep}}, x_\text{0,ego}, v_\text{wind}, \mu_\text{road}]$ (Figure~\ref{fig:6d}). Note that the wind and road parameters are not explicitly considered in the modeling and closed-loop control of the developed motion planner. Therefore, they should act as disturbance factors or modeling errors. The parameter ranges of the logical scenario, $S$, are enlisted in Table~\ref{table_sym}.\\

\begin{table}[H]
  \centering
  \begin{tabular}{ccccc}
    \toprule
    \textbf{Parameter description} & \textbf{symbol} & \textbf{Unit} & \textbf{range} \\
    \midrule
    Ego velocity & $v_{\text{ego}}$ & [m/s] & [5,25] \\
    \midrule
    Leading vehicle velocity & $v_{\text{lead}}$ & [m/s] & [0,20] \\
    \midrule
    Initial relative distance  & $d_{\text{0,sep}}$ & [m] & [50,100] \\
    \midrule
    Initial lateral ego position & $x_\text{0,ego}$ & [m] & [-0.5,0.5]] \\
    \midrule
    Wind gust velocity & $v_\text{wind}$ & [m/s] & [0,25]  \\
    \midrule
    Road friction & $\mu_\text{road}$ & [-] & [0.2,0.8] \\
    \bottomrule
  \end{tabular}
   \caption{Parameters and their range for the logical scenario description}
     \label{table_sym}
\end{table}

We compare the performance of our framework with simulations performed using full factorial combinatorial testing. For the baseline simulations, only the blue arrow loop of Figure~\ref{fig:arch} is followed where the seed sample set, $S_\text{seed}$ is a combination of discretized values considered parameter ranges included in the logical scenario, covering the parameter space uniformly. Instead, for the BO framework, the seed sample set is chosen based on the number of parameters involved according to the research in~\cite{frazier_tut}. We use the rule of thumb of 5x seed simulation runs for training the surrogate model. Additional tests of the green-arrow Bayesian optimization loop in Figure~\ref{fig:arch} are then performed until the stopping criteria are met.\\

The co-simulation environment couples a model-predictive controller in MATLAB/Simulink with the IPG CarMaker vehicle simulator for closed-loop behavior~\cite{MATLAB,IPG}. In Simulink, the ego vehicle executes a rule-defined motion planner based on an \emph{adaptive MPC} framework, formulated via CasADi and solved by \texttt{acados}~\cite{casadi,acados}. Following the work~\cite{Rule_def_MPC} and~\cite{MPC2}, this planner embeds high-level driving rules, which include overtaking, following or stopping, directly within the MPC by adapting both the cost-function weights and \ac{CBF} parameters (\(\gamma\) and \(D_{\rm OV}\)) based on the scenario detected. The weights remain unchanged from the work presented in~\cite{MPC2} and the correct sets of weights are automatically chosen during each simulation run. For instance, during overtaking, lateral deviation is penalized less while velocity tracking is emphasized, paired with a tight safety margin; conversely, during following or stopping, the longitudinal error is weighted more heavily and the CBF parameters are adjusted to prioritize lane adherence or complete stop. The obstacle vehicle follows a simple velocity profile. The Simulink MPC outputs acceleration or brake and steering commands to the simulator, which models realistic vehicle dynamics. This closed-loop setup captures how control decisions (e.g., steering to avoid the obstacle) interact with the vehicle physics.  For each concrete scenario $x$, the co-simulation runs for a fixed duration or until an off-road event. Note that no collisions occur because of the inherent modeling of the MPC where \ac{CBF} is a hard constraint. Two failure modes exist. The first one includes off-road events due to runtime solver failure, in which the solver initially finds a solution but eventually fails due to the constraints present. The second failure mode is the \textit{a priori} solver infeasibility, failing to find any valid solution to the simulation requested.\\

Although standard criticality metrics such as Time-to-Collision (TTC) or Post-Encroachment Time (PET) could serve as the objective function~\cite{criticality_metrics}, this work uses two more direct metrics: the maximum lateral displacement of the ego vehicle and the solver infeasibility status. This choice is motivated by the two failure modes under consideration as discussed above. Off-road events can occur when the ego vehicle, influenced by the Control Barrier Function (CBF) of the leading vehicle, overshoots the road boundary. Minimizing lateral displacement, therefore, directly addresses this specific failure mechanism. The solver status is instead  a discrete metric that indicates whether the infeasibility  of the solution is due to the initial conditions (the scenarios is unable to be handled by the solver) or eventual runtime infeasibility due to the MPC constraints.

The iteration of the Bayesian Optimization framework simulations terminate when at least one instance in each of the critical clusters identified in the baseline simulations are identified. The identification of clusters for the given case study is discussed more in detail in Section~\ref{sec:5}. However, an upper bound for the number of simulations to be performed is set at one order of simulations less than the ones presented in the baseline case. This has been done to analyze the performance of the validation framework as the framework would not be beneficial if the simulations required are in the same order as those of the baseline case. Different combinations of the acquisition functions and  Hence this paper also benchmarks for the stopping criteria by comparing the baseline simulations with the ones executed with the BO framework. The statistical analysis of the results obtained using different combinations of acquisition functions and criticality metrics are present in Section~\ref{sec:5}.

%% file: Sections_V1/S5_Results.tex
\section{Case Study Results}
\label{sec:5}
\subsection{Case A: 3-DoF simulations}
\label{subsec:3dof}
\begin{figure}[!t]
  \centering
  \includegraphics[width=\columnwidth]{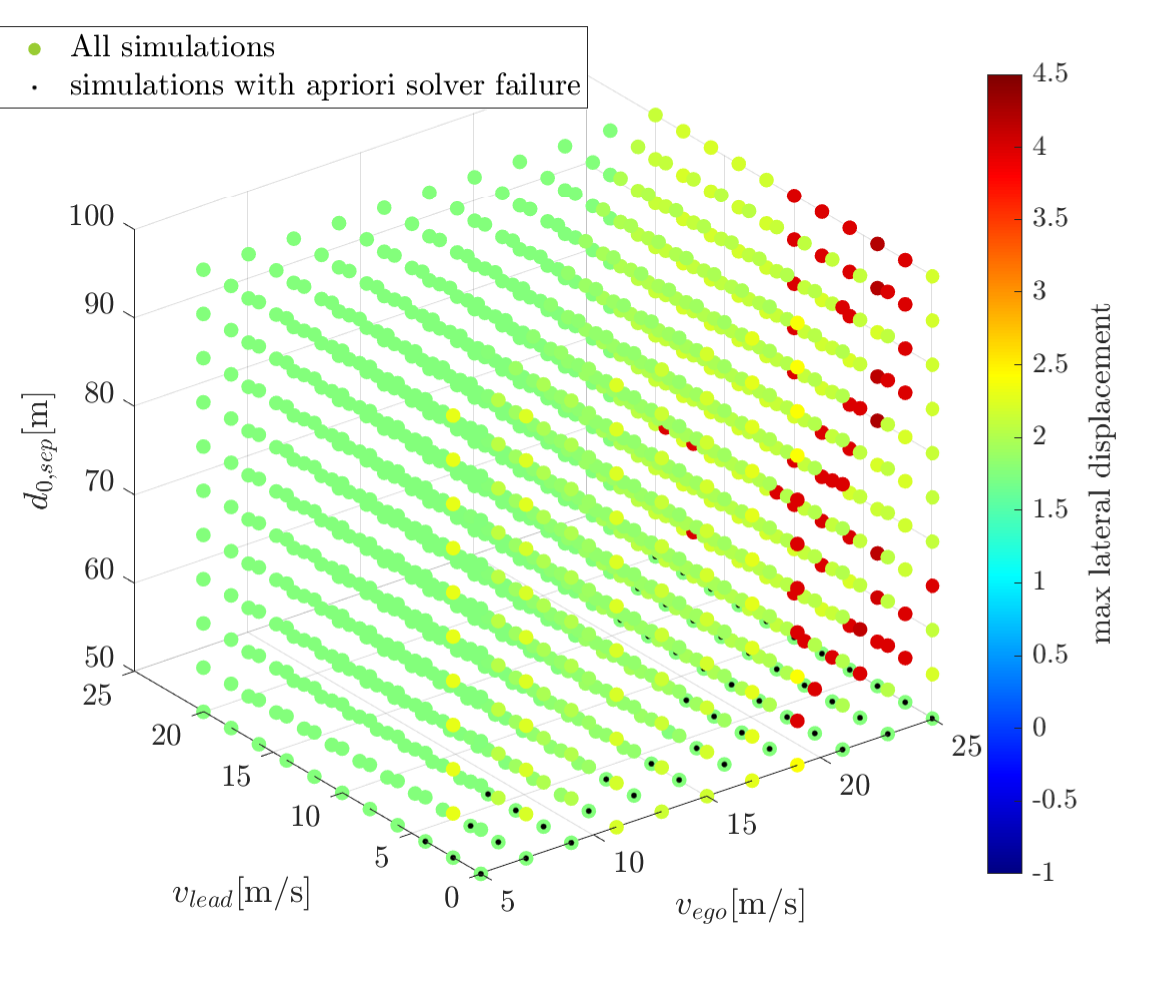}
  \caption{Baseline simulation results of the 3-DOF simulations. The parameter sweep covers ego speed, obstacle speed, and initial distance between them. Off-road scenarios are marked in red, representing critical failures.}
  \label{fig:baseline_gt_3dof}
\end{figure}

\begin{figure}[!t]
  \centering
  \includegraphics[width=\columnwidth]{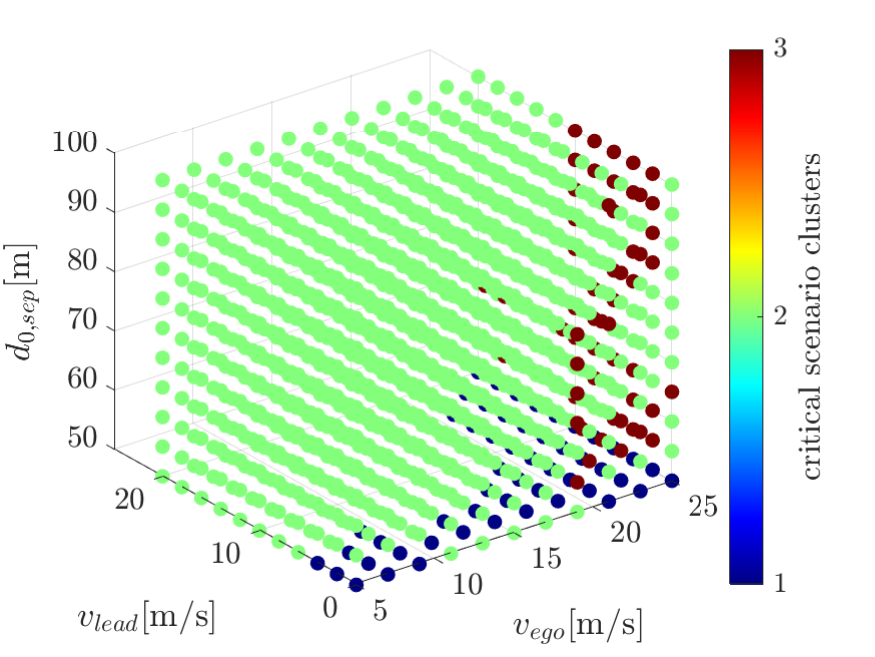}
  \caption{The two principal clusters identified by the two criticality metrics used: A continuous metric-max lateral displacement of the \ac{AV} and a discrete metric-MPC solver status}
  \label{fig:clusters}
\end{figure}
To establish a baseline for comparison, an exhaustive full-factorial Design of Experiments (DoE) was conducted across a discretized, three-degree-of-freedom (3-DOF) parameter space. This 3-DOF case was chosen because it represents the maximum dimensionality in which the results can be completely visualized, thereby providing an intuitive understanding of the methodology. The scenario parameters include the ego vehicle's speed (ranging from $5$ to $25~\text{m/s}$ in $2~\text{m/s}$ increments), the leading vehicle's speed (from $0$ to $20~\text{m/s}$ in $2~\text{m/s}$ increments), and the initial spacing (from $50$ to $100~\text{m}$ in $5~\text{m}$ increments). This discretization results in $11 \times 11 \times 11 = 1331$ unique scenarios.

For each configuration, a co-simulation evaluates the ego vehicle's trajectory under MPC control. As discussed in Section~\ref{sec:4}, the motion planner can exhibit two types of failures; consequently, this case study uses two corresponding criticality metrics: one continuous and one discrete. The continuous metric is the maximum lateral displacement of the vehicle from the centerline, with a failure threshold set to $3.5~\text{m}$ to match the lane width. The discrete metric is the solver's infeasibility status, indicating the operational limits of the controller, where a status of $4$ denotes a null simulation (due to \textit{a priori} infeasibility) and $0$ represents a successfully executed simulation. A single value for all the criticality metrics considered is obtained from each simulation run.\\
Figure~\ref{fig:baseline_gt_3dof} summarizes these results, highlighting the value of the continuous criticality metric---the absolute maximum lateral displacement from the centerline. As shown in the figure, the most extreme cases correspond to combinations of high relative speed and small initial spacing between the vehicles. It is also evident that failures (off-road events) occur in the region where this metric exceeds the $3.5~\text{m}$ lane-width threshold.
Simulations resulting in \textit{a priori} failure due to initial condition infeasibility are marked with a dot inside the simulation marker in Figure~\ref{fig:baseline_gt_3dof}. In these cases, since no trajectory data is produced, the maximum lateral displacement metric equals the centerline of the lane, that is $1.75~\text{m}$, which is the initial starting position of the ego vehicle. This comprehensive dataset serves as the ground truth for validating the sample efficiency of the Bayesian optimization approach, which seeks to discover these same critical regions using significantly fewer simulations. \\
\begin{figure*}[t] 
  \centering

  \begin{subfigure}[b]{0.45\textwidth}
    \centering
    \includegraphics[width=\textwidth]{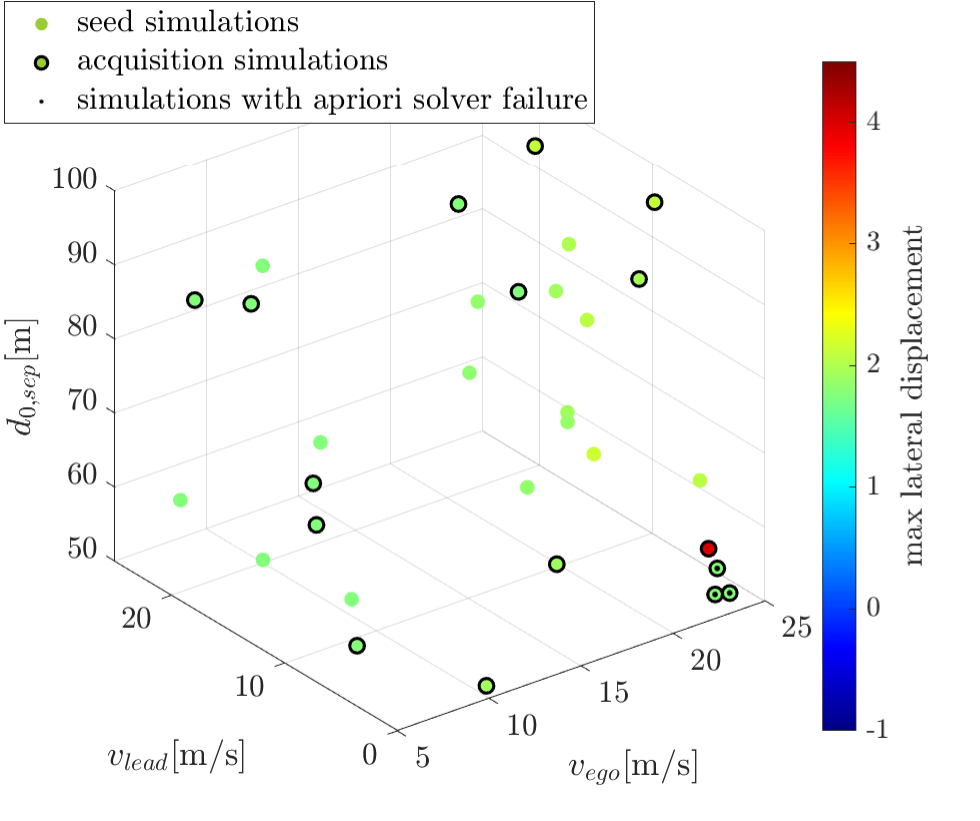}
    \caption{Simulation A: Thompson Sampling and MPC Solver status}
    \label{fig:simA}
  \end{subfigure}
  \hfill
  \begin{subfigure}[b]{0.45\textwidth}
    \centering
    \includegraphics[width=\textwidth]{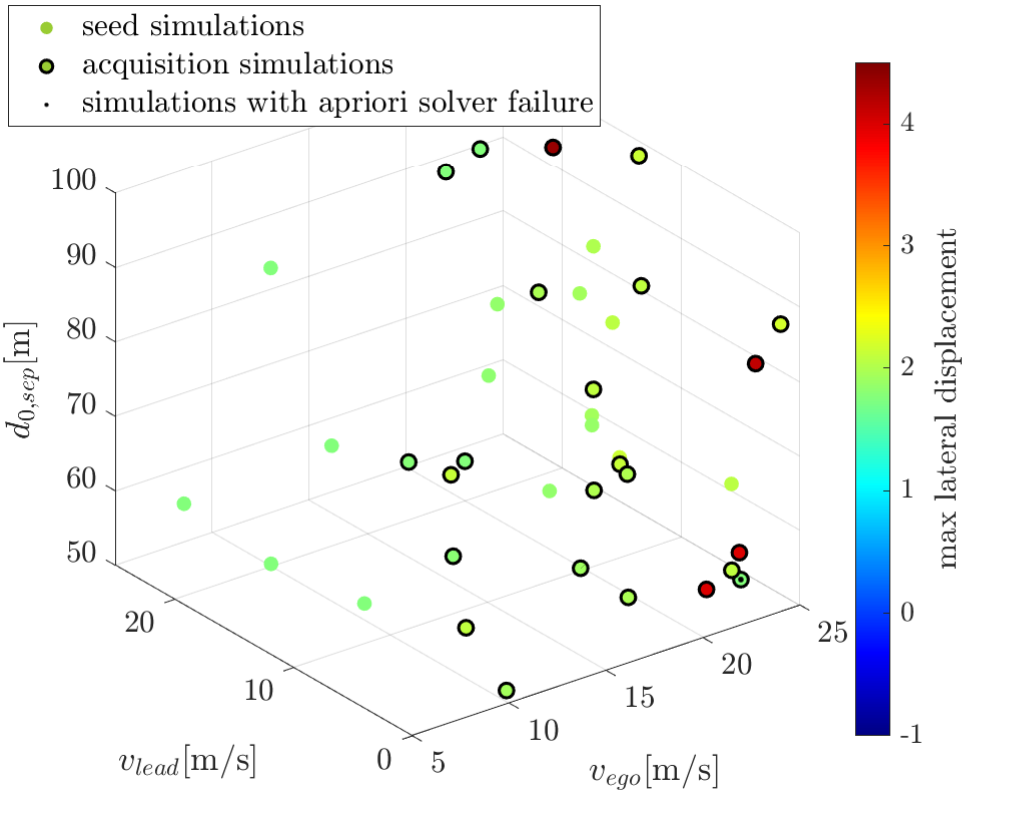}
    \caption{Simulation B: Thompson Sampling and max lateral displacement}
    \label{fig:simB}
  \end{subfigure}
  \hfill
  \begin{subfigure}[b]{0.45\textwidth}
    \centering
    \includegraphics[width=\textwidth]{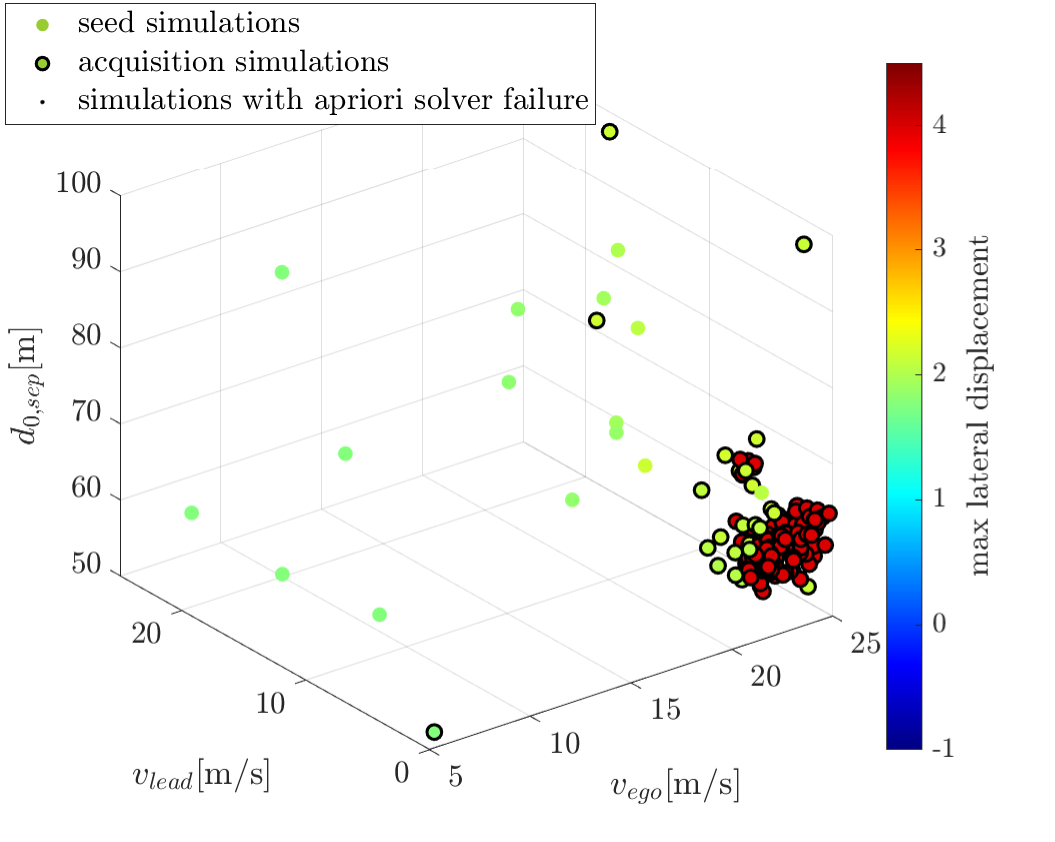}
    \caption{Simulation C: Probability of Improvement and max lateral displacement}
    \label{fig:simC}
  \end{subfigure}
  \hfill
  \begin{subfigure}[b]{0.45\textwidth}
    \centering
    \includegraphics[width=\textwidth]{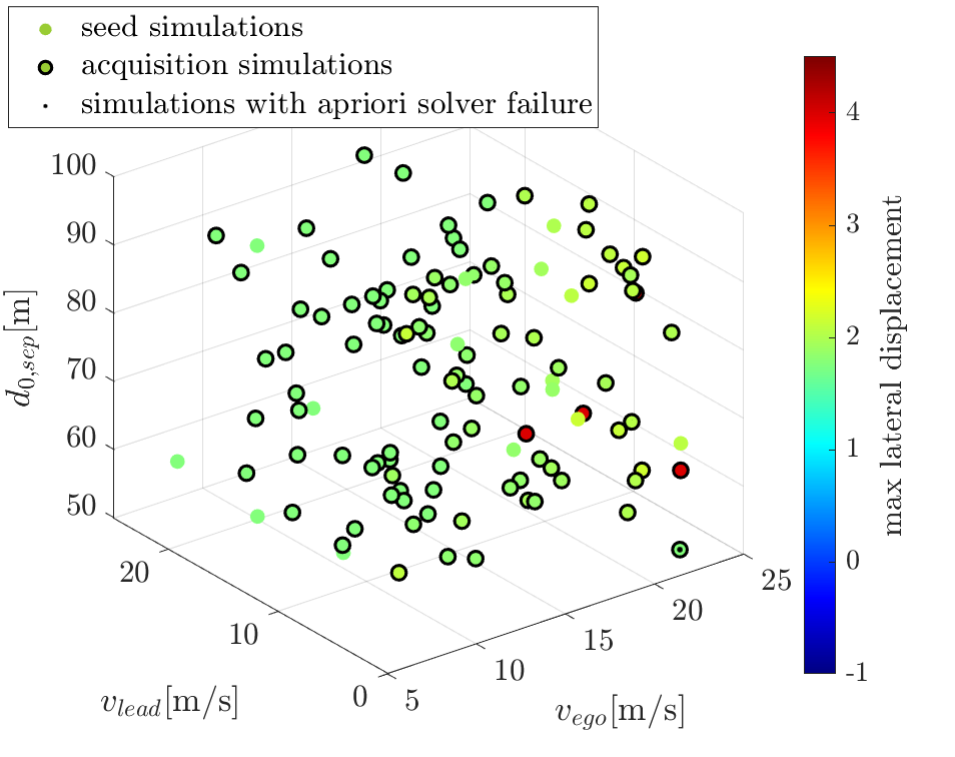}
    \caption{Simulation D: Probability of Improvement and MPC Solver status}
    \label{fig:simD}
  \end{subfigure}

  \caption{One complete run each of the 3 simulation categories with the same seed sample. The seed runs and the acquisition runs are distinguished by the marker edge: (a) discrete metric triggers exploration of the parameter space,(b) continuous metric leads to exploitation of overtake scenario parameters (c) exploitation of local minimum, only one cluster identified (d) exploration of the full logical scenario space until both critical types are found.}
  \label{fig:threeimages}
\end{figure*}
To identify the critical zones in the logical scenario space, clustering of the criticality metrics resulting from the baseline simulations is performed. \ac{DBSCAN} algorithm was used to cluster the results into groups of varying sizes~\cite{dbscan}. This clustering algorithm was chosen so that it could be applied to any $n$-dimensional space and hence it could account for multiple criticality metrics as input. Figure~\ref{fig:clusters} shows the three clusters found using the criticality metric vector as input to the algorithm. Two of these three clusters are critical ones pertaining to the ego crossing the road bounds (Cluster 3) and \textit{a priori} solver failure (Cluster 1). The third and the biggest cluster (Cluster 2) is of the simulations where the controller successfully performs the necessary manoeuvre. The size of the clusters is reported in Table~\ref{table_clusters}.  \\
\begin{table}[H]
  \centering
  \begin{tabular}{ccccc}
    \toprule
    \textbf{Simulation set} & \textbf{Cluster 1} & \textbf{ Cluster 2} & \textbf{Cluster 3} & \textbf{Total simulations} \\
    \midrule
    3-DoF & 52 & 1223 & 56 & \textbf{1331} \\
    \midrule
    6-DoF & 720 & 12906 & 6174 & \textbf{19800} \\
    \bottomrule
  \end{tabular}
   \caption{Sizes of the clusters found in the baseline simulations. Clusters 1 and 3 are the critical clusters related to off-road events and \textit{a priori} solver failure respectively.}
     \label{table_clusters}
\end{table}
The validation of the Bayesian Optimization (BO) framework aims to confirm its ability to rapidly explore and identify critical scenarios. While a single detected failure would typically invalidate the system for a fixed Operational Design Domain (ODD), one of the objectives of this work is to analyze the capability of the framework to detect multiple, distinct failures or critical zones. To achieve this objective, multiple acquisition functions as previously mentioned in Section~\ref{sec:3} have been thoroughly tested. The choice of acquisition function can balance the exploration with exploitation and can prevent the optimization getting stuck in a local minimum. Results including two of these acquisition functions, Thompson Sampling and Probability of Improvement(PI), in the test runs have been presented henceforth. Thompson Sampling naturally balances exploration and exploitation whereas PI tends to be greedy and more exploitative.. 
\\
Four sets (A, B, C and D) of 10 simulations each were performed to study the behaviour of using different types of criticality metrics (continuous vs discrete) to form the surrogate model and different acquisition functions (balanced vs greedy) for evaluating the next concrete scenario simulation parameter values. 
One instance each of these 4 sets with the same seed samples for training the surrogate model is shown in Figure~\ref{fig:threeimages}. Simulation sets A and B both use Thompson Sampling as the acquisition function whereas C and D use Probability of Improvement (PI) function. Furthermore, A and D use the discrete criticality metric (solver status) while B and C use the continuous metric (maximum lateral displacement of the ego).  It is apparent from Figure~\ref{fig:simA} that since the surrogate is trained on the discrete metric, the acquisition simulations are based on pure exploration of the parameter ranges until the critical cases are found in the lower right corner whereas in Figure~\ref{fig:simB} the acquisition simulations are biased towards the higher velocity difference between the two vehicles where the overtaking and hence the higher lateral displacement occurs. In Figure~\ref{fig:simC}, the exploitation is more prevalent than exploration and once the failure is detected the new acquisition points remain close by. From Figure~\ref{fig:simD}, exploration characteristic of the PI acquisition function is visualized since using a discrete metric does not add any new information to the surrogate model until the critical zone is found by searching in uncertain areas of the parameter space. Furthermore, in the plotted case of simulations C, critical scenarios pertaining to Cluster 1 were not found while in A, B, and D, instances in both clusters were successfully identified. \\ 
\begin{figure}[!t]
  \centering
  \includegraphics[width=\columnwidth]{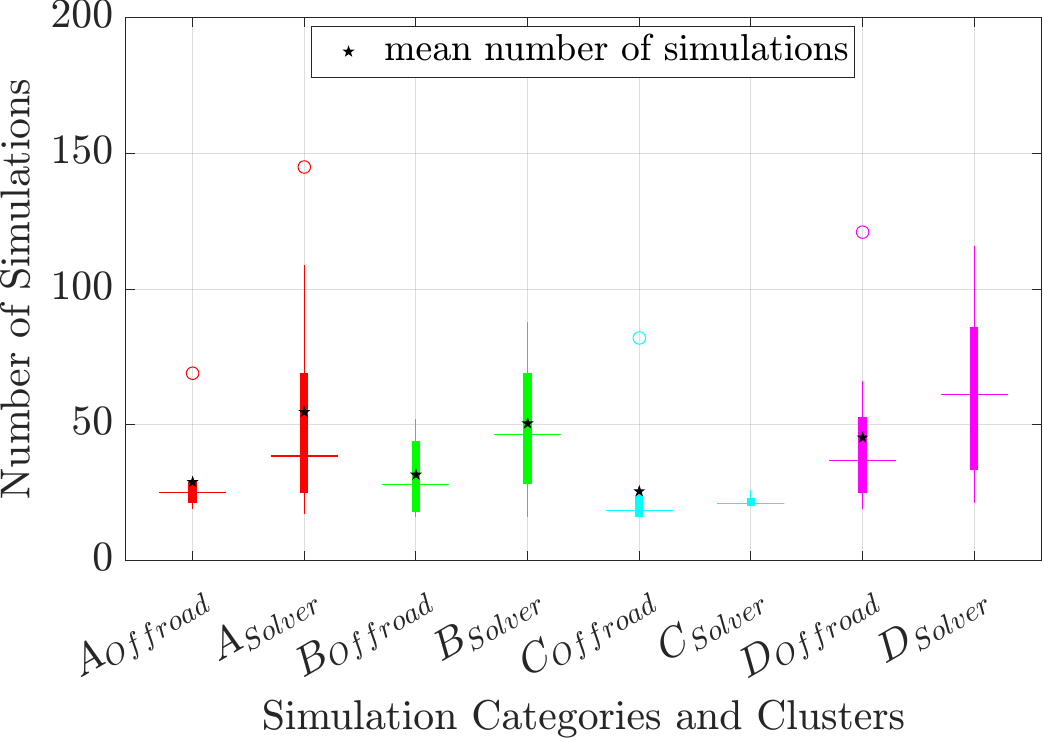}
  \caption{4 categories of 10 sets of simulations. A: Criticality metric is solver status(discrete) and the acquisition function is Thompson sampling. B: Criticality metric is max lateral displacement of ego(continuous)and the acquisition function is Thompson sampling. C: Criticality metric is max lateral displacement of ego(continuous)and the acquisition function is Probability of Improvement. D: Criticality metric is solver status(discrete) and the acquisition function is Probability of Improvement.}
  \label{fig:cluster_BO}
\end{figure}
The box plots of the results that compare these simulations is presented in Figure~\ref{fig:cluster_BO}. The plots present a statistical analysis of the number of simulations performed before the first detection of critical scenario in each identified critical cluster starting with different seed simulations. This difference occurs due to the stochastic Latin Hypercube Sampling of the initial simulations. The numbers include the initial seed simulation runs, which is equal to 15 in this case. 
The upper bound for the framework iterations was set at 150 simulations. It can be seen from the box plot that the mean and median number of simulations required in each case to find the first critical scenario of each cluster is less than 100 simulations which is one order less than the baseline simulations. However, Simulation C and D, using the PI acquisition function were unable to find both critical zones 5 and 2 times out of 10 respectively, hence the mean values in this case haven't been included. It can also be observed that the discovery of the offroad events' cluster has less variance and gets detected earlier in average when compared to the solver-failure cluster. The median simulations required to find the solver-failure cluster is always greater but always remain in the same order less than the baseline simulations. Thus, from Figure~\ref{fig:threeimages} and Figure~\ref{fig:cluster_BO} it can be seen that continuous metrics drive the discovery of critical points  by creating meaningful surrogate model whereas discrete metrics rely on the exploration of the scenario space by the acquisition function to do so. If exploitation-favored configurations of acquisition functions like PI are used with other metrics (the case of Simulation C), it's difficult to discover critical instances pertaining to the discrete metric using this framework. However, using acquisition functions that are inherently balanced between exploration and exploitation like Thompson Sampling may solve this issue although requiring more simulations in average. The combination of using Thompson Sampling with a continuous metric (Simulation B) is the only one without outliers.
\subsection{Case B: 6-DoF simulations}
\begin{figure*}[!t]
  \centering
  \includegraphics[width=\textwidth]{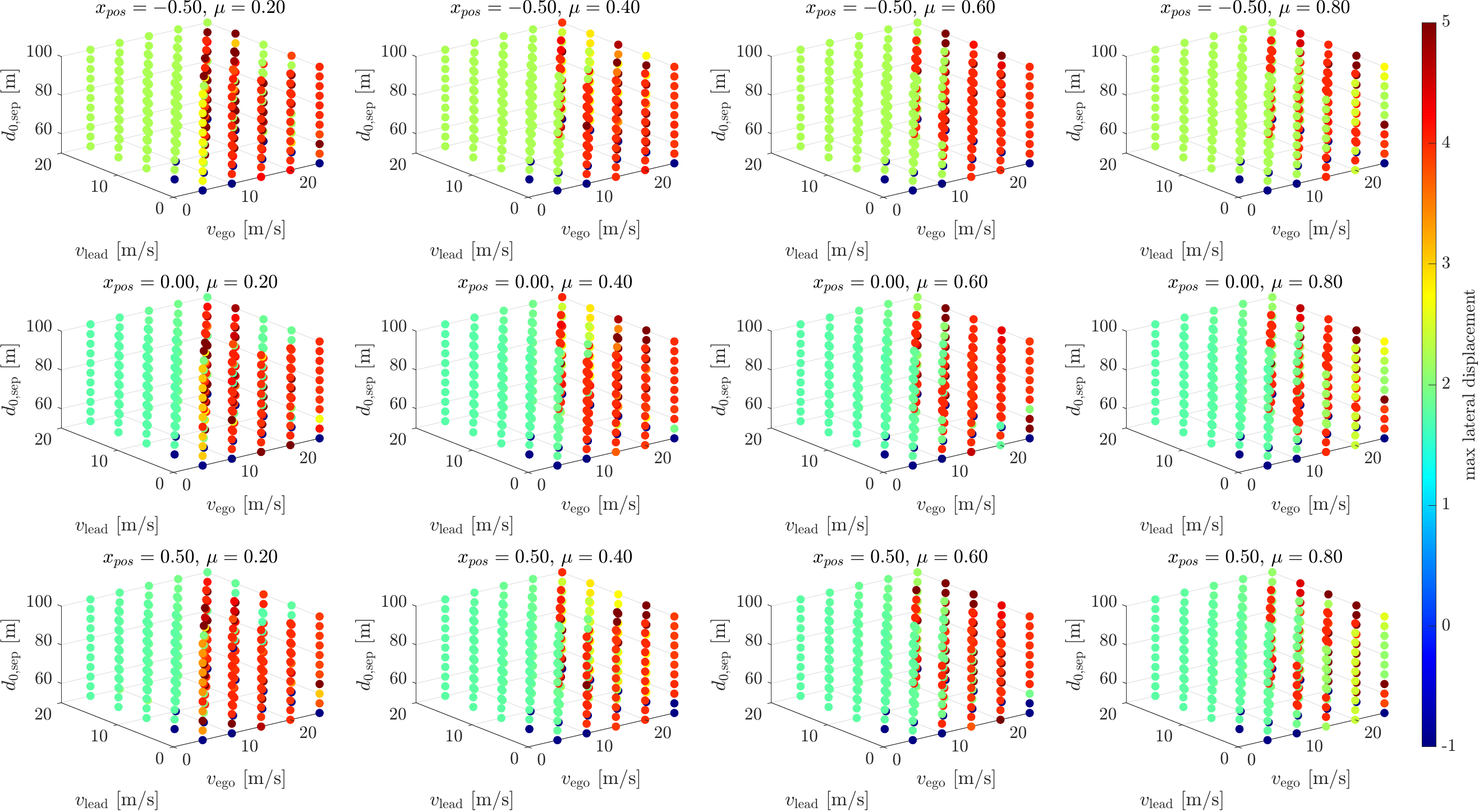}
  \caption{Simulations performed for wind gust velocity = 15 m/s. each subplot represents the 3-D equivalent simulations for each discrete value combination of the road friction coefficient and starting x position along the road.}
  \label{fig:GT_6D}
\end{figure*}
To evaluate the framework's scalability, the parameter space was expanded to six dimensions by adding three new parameters, with ranges as defined in Section~\ref{sec:4}. These new parameters are the ego vehicle's initial lateral offset from the centerline, the lateral wind gust velocity, and the road friction coefficient. The wind velocity was varied up to 25 m/s, corresponding to a 'whole gale' on the Beaufort scale (Force 10). The road friction coefficient, $\mu$, ranged from 0.2 (representing icy conditions) to 0.8 (representing dry asphalt).
In total, $5 \times 5 \times 11 \times 3 \times 6 \times 4 =  19,800$ baseline simulations were performed across the expanded parameter space as presented in Table~\ref{table_sym}. A representative subset of these results, for a constant wind velocity of 15 m/s, is shown in Figure~\ref{fig:GT_6D}. The critical scenarios are marked in red, representing crossing of the road limits. The \textit{a priori} solver failure simulations are represented as dark blue dots by assigning them a value of $-1$ for ease of visualization only. It could be seen from the subplots that wind and low road frictions have a destabilizing effect on the vehicle and the aggressive motion-planner commands at high speed overtake maneuvers result in the vehicle exiting the road boundaries almost throughout the set ODD. A decrease in the number of critical scenarios from the leftmost to the rightmost subplots of the image can also be observed. This is due to the increase in road friction parameters as reported in the figure. The relative starting position in the given range has little effect on the criticality of the simulation as the subplots remain almost the same from the top to the bottom.
\begin{figure}[!t]
  \centering
  \includegraphics[width=\columnwidth]{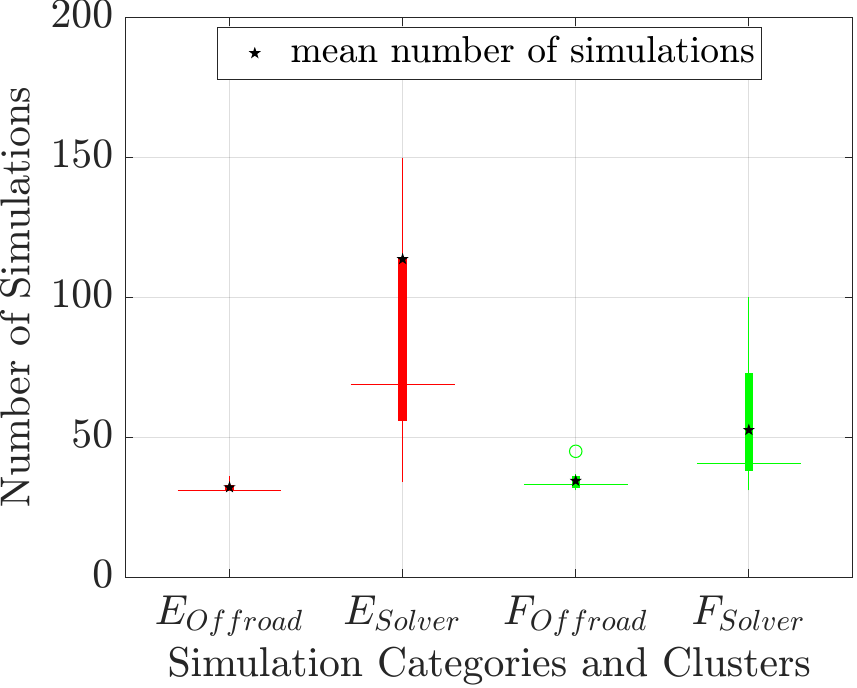}
  \caption{2 categories of 10 sets of simulations. A: Criticality metric is max lateral displacement of ego(continuous) and the acquisition function is Thompson sampling. B: Alternate criticality metrics are used to form the surrogate: max lateral displacement of ego(continuous) and the solver status(discrete). The acquisition function is Thompson sampling.}
  \label{fig:cluster_BO6}
\end{figure}
This 6-DOF case involves an order of magnitude more simulations than the 3-DOF analysis presented previously. Using the DBSCAN clustering algorithm, two distinct critical clusters were again identified, corresponding to \textit{a priori} solver failure and off-road events, consistent with the findings in Section~\ref{subsec:3dof}. The sizes of these clusters are detailed in Table~\ref{table_clusters}. The objective remains to evaluate the \ac{BO} framework's capability to efficiently detect these distinct critical zones.

To assess the framework's performance, two simulation sets (E and F) were conducted, with results summarized in Figure~\ref{fig:cluster_BO6}. For both sets, the initial seed sample size was set to five times the parameter space dimension, amounting to 30 scenarios. To ensure a valid comparison, these 30 initial simulations are included in the total counts, and the same seed samples were used for both Set E and Set F.

Simulation Set E is analogous to the best-performing approach from the 3-DOF case, where the continuous metric (maximum lateral displacement) was used to train the surrogate model, and Thompson Sampling selected subsequent test cases. This approach proved highly efficient at discovering the off-road events, likely because the large size of this cluster increased the probability of finding a critical point within the initial seed set. However, this single-metric strategy struggled to identify the smaller solver-failure cluster, exhibiting high variance and requiring significantly more simulations. The mean number of simulations for this task was skewed by an outlier run of 472 simulations (not shown in the figure), highlighting the potential unreliability of relying on a single metric to find all failure types.

To address this limitation, a hybrid strategy was tested in Set F. This approach alternates between the two criticality metrics obtained as the results of simulations to train the surrogate model in subsequent iterations. As shown in Figure~\ref{fig:cluster_BO6}, the hybrid strategy of Set F yields a superior and more balanced performance. Training the surrogate on the discrete solver-failure metric promotes broader exploration of the parameter space, which significantly reduces the mean, median, and variance of the simulations required to identify the smaller, harder-to-find cluster. The framework's overall efficiency remains comparable to the 3-DOF case, indicating that it is highly scalable. In this 6-DOF study, the proposed method is two orders of magnitude more efficient than the baseline combinatorial testing.

%% file: Sections_V1/S6_Conclusions_n_FutureWork.tex
\section{Conclusion and Future Work}
\label{sec:6}
This paper introduces a framework for the accelerated virtual validation of an \acf{ADF}. The framework combines Scenario-Based Testing (SBT) with Bayesian Optimization to efficiently discover critical scenarios by iteratively fitting a Gaussian Process surrogate model to simulation outcomes. New test cases are selected using acquisition functions that maximize a predicted failure metric. This approach identifies safety violations, such as off-road events and solver failures, with orders of magnitude fewer simulations than brute-force methods. A key advantage of this work is its flexible, model-agnostic workflow, which can handle diverse logical scenario definitions and high-dimensional parameter spaces within a given \ac{ODD}.

The framework's effectiveness was demonstrated through a use case involving a \acf{MPC}-based motion planner operating on a two-lane highway with preceding traffic. A co-simulation platform integrating MATLAB/Simulink, the acados MPC solver, and IPG CarMaker provided a realistic testbed. The results show that the framework successfully identifies distinct critical clusters that were previously found via baseline full-factorial testing. The method performs most robustly when the surrogate model is developed with continuous criticality metrics and when using acquisition functions, such as Thompson Sampling, that naturally balance exploration and exploitation. Furthermore, a hybrid strategy that alternates between different criticality metrics to train the surrogate model was explored and found to be highly effective.

To demonstrate scalability, the framework was applied to the same \acf{ADF} while doubling the parameter space from three to six variables. The approach proved to be scalable, detecting the critical clusters up to two orders of magnitude more efficiently than the baseline ground-truth simulations. This framework, therefore, shows significant promise for expediting the development and homologation processes of autonomous driving functions when integrated into virtual testing toolchains.

Future work will focus on applying this framework to more complex, multi-agent scenarios in urban environments and integrating perception and localization modules into the simulation loop. Further extensions will explore multi-fidelity modeling and multi-objective optimization to simultaneously find failures under different criteria. Ultimately, this line of research aims to provide rigorous and scalable tools for \ac{AV} safety assurance by focusing testing efforts on the most challenging scenarios.
